\documentclass[runningheads]{llncs}
\usepackage{amsmath, amsfonts, graphicx}
\usepackage{CJK}
\usepackage{graphicx,subfigure}
\makeatletter
\newcommand{\printfnsymbol}[1]{%
  \textsuperscript{\@fnsymbol{#1}}%
}

\usepackage{booktabs}

\usepackage{amssymb}
\usepackage{pifont}
\newcommand{\cmark}{\ding{51}}%
\newcommand{\xmark}{\ding{55}}%

\makeatother
\usepackage{hyperref,xcolor}
\hypersetup{hypertex=true,
            colorlinks=true,
            linkcolor=red,
            anchorcolor=blue,
            citecolor=green,
            urlcolor=blue
            }

\begin{document}
\title{Multi-Modal Representation Learning with Self-Adaptive Threshold for Commodity Verification}
\titlerunning{Multi-Modal Representation Learning with SAT for Commodity Verification}
%
\author{Chenchen Han \orcidID{0000-0002-3330-5308}\thanks{Equal contribution. Listing order is random.} \and
Heng Jia \orcidID{0000-0001-8062-9455}\printfnsymbol{1}}
\authorrunning{C. Han and H. Jia}
%
\institute{Zhejiang University, Zhejiang, China \\
\email{hanchenchen@zju.edu.cn  jiaheng.dlut@gmail.com }\\
}
\maketitle              
\begin{abstract}
In this paper, we propose a method to identify identical commodities. In e-commerce scenarios, commodities are usually described by both images and text. By definition, identical commodities are those that have identical key attributes and are cognitively identical to consumers. 
There are two main challenges: 1) The extraction and fusion of multi-modal representation. 2) The ability to verify identical commodities by comparing the similarity between representations and a threshold. To address the above problems, we propose an end-to-end multi-modal representation learning method with self-adaptive threshold. We use a dual-stream network to extract multi-modal commodity embeddings and threshold embeddings separately and then concatenate them to obtain commodity representation. Our method is able to adaptively adjust the threshold according to different commodities while maintaining the indexability of the commodity representation space.
We experimentally validate the advantages of self-adaptive threshold and the effectiveness of multimodal representation fusion. Besides, our method achieves third place with an F1 score of 0.8936 on the second task of the CCKS-2022 Knowledge Graph Evaluation for Digital Commerce Competition. Code and pretrained models are available at \url{https://github.com/hanchenchen/CCKS2022-track2-solution}.

\end{abstract}
\section{Introduction} 

We aims to identify identical commodities based on representation learning. Given a pair of commodities, we extract their representations and calculate the similarity between representations. Then we judge whether the pair is identical by comparing the similarity and threshold. In the second task of the CCKS-2022 Knowledge Graph Evaluation for Digital Commerce Competition, the commodity pair data is from the recall results of actual online models and manually labeled, where most of the negative pairs are similar but some key attributes do not match. 

The traditional identical commodity verification methods usually adopt manually adjusted thresholds. There are some disadvantages of such a method.

1) \textbf{Inter-dataset adaptation problem}. Since data distribution usually varies between datasets, the corresponding representation distribution will be different as well. The threshold determined on one dataset may be hard to achieve comparable results on another, which affects the generalization of the model. It is necessary to manually adjust the threshold, which is laborious and burdensome. 

2) \textbf{Intra-dataset adaptation problem}. Since the commodity pairs are usually similar, there representations often crowded together in the representation space. A slight fluctuation of the threshold may affect the performance much. Moreover, it is unwise to use the same threshold for different kinds of commodities.

3) \textbf{Model optimization problem}. Due to the high similarity of commodities, their similarity scores are usually higher than 0. However, the existing loss functions (e.g., binary cross entropy loss) are usually centered at 0. Consequently, the model is difficult to be optimized. Besides, it may destroy the representation space to force pushing the representations of similar but non-identical commodities away.

To mitigate the above problems, we propose an end-to-end multi-modal representation learning method with Self-Adaptive Threshold (SAT). We use a dual-stream network to extract multi-modal commodity embeddings and threshold embeddings separately and then concatenate them to obtain commodity representation. Our method can adaptively adjust the threshold according to different commodities, thus reducing the burden and drawbacks of manually adjusting the threshold. 
The dual-stream network optimizes the commodity representation distribution bidirectionally by either the commodity stream or the threshold stream, which results in a better distribution of representations. Therefore, it is less likely to force pushing away the representations of similar but different commodities. Moreover, with our self-adaptive threshold, the similarity of representations is basically centered at 0. While maintaining the indexability of the commodity representation space, the model is easier to be optimized and the representations are more robust (more details in Section \ref{distribution}).

Our main contributions are as follows:

1) We analyze the possible problems in the traditional commodity verification approach and then propose a multi-modal representation approach with SAT to learn the threshold adaptively. Our approach reduces the burden of adjusting thresholds and enhances the generalization and robustness of the representations. 

2) We do not do special processing for the inputs (e.g. no detector), and the whole network is trained end-to-end so that other methods can be easily integrated.

3) We experimentally validate the advantages of the self-adaptive threshold and the effectiveness of our multi-modal representation fusion. Our method achieves an F1 score of 0.8936 and takes third place on the second task of the CCKS-2022 Knowledge Graph Evaluation for Digital Commerce Competition.


\begin{figure}
    \centering
    \includegraphics[width=1.0\textwidth]{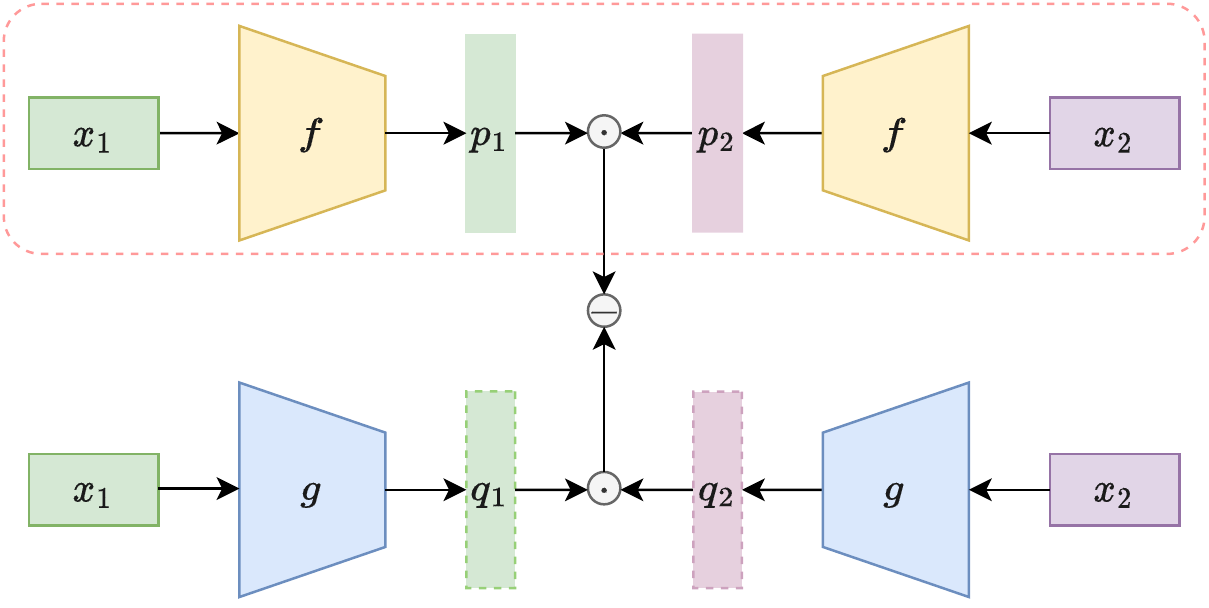}
    \caption{
    Multimodal representation learning with self-adaptive threshold. The area in the red box is the traditional method of calculating the similarity of a commodity pair. The score is obtained by subtracting the pre-defined threshold from the similarity. When the score is greater than zero, the commodity pair is predicted to be identical, and vice versa. The higher score, the greater probability of being an identical commodity pair. We add a threshold stream to learn self-adaptive threshold embeddings and regard the difference between the inner product of commodity embeddings and threshold embeddings as the score.
    }
    \label{fig:model}
\end{figure}

\section{Method}

In this section, we present SAT, a novel multi-modal representation learning method with self-adaptive threshold for commodity verification. We first detail the self-adaptive threshold in Sec. \ref{sat}, then introduce the model architecture in Sec. \ref{model} and the loss function in Sec. \ref{loss} finally. Fig. \ref{fig:model} shows the overview of the proposed method.

\subsection{Self-Adaptive Threshold}\label{sat}

\subsubsection{Dual-Stream Embedding} We propose to use a dual-stream network to extract the commodity embedding and threshold embedding. Given a commodity $\boldsymbol{x}$, we feed it to the commodity-stream $\boldsymbol{f}$, and extract commodity embedding:

\begin{equation}
    \boldsymbol{p} = \boldsymbol{f}(\boldsymbol{x})
\end{equation}

where $\boldsymbol{p} \in \mathbb{R}^{d1}$ is the commodity embedding; $d1$ is the commodity embedding dimension. Correspondingly, we have a threshold-stream $\boldsymbol{g}$ to extract the threshold embedding:

\begin{equation}
    \boldsymbol{q} = \boldsymbol{g}(\boldsymbol{x})
\end{equation}

where $\boldsymbol{q} \in \mathbb{R}^{d2}$ is the threshold embedding; $d2$ is the threshold embedding dimension. Then we can acquire the complete embedding of commodity $\boldsymbol{x}$ by concatenation:

\begin{equation}
    \boldsymbol{z} = [\boldsymbol{p}, \boldsymbol{q}]
\end{equation}

where $\boldsymbol{z} \in \mathbb{R}^{d1+d2}$ is the complete embedding; $d1 + d2$ is the embedding dimension; $[\cdot, \cdot]$ represents concatenation.

\subsubsection{Score Calculation}

As our method is based on representation learning, we do not have to tackle a commodity pair simultaneously. Given a commodity pair $(\boldsymbol{x_{1}}, \boldsymbol{x_{2}})$, we extract their embeddings separately:

\begin{equation}
    \begin{split}
        \boldsymbol{z_{1}} = [\boldsymbol{p_{1}}, \boldsymbol{q_{1}}] \\
        \boldsymbol{z_{2}} = [\boldsymbol{p_{2}}, \boldsymbol{q_{2}}] \\
    \end{split}
\end{equation}

The similarity $s$ is obtained by the inner product between commodity embeddings $\boldsymbol{p_{1}}$ and $\boldsymbol{p_{2}}$:

\begin{equation}
    s = \boldsymbol{p_{1}} \cdot \boldsymbol{p_{2}}
\end{equation}

where $\cdot$ represents the inner product between vectors. Correspondingly, we can get the self-adaptive threshold by the inner product between threshold embeddings $\boldsymbol{q_{1}}$ and $\boldsymbol{q_{2}}$:

\begin{equation}
    t = \boldsymbol{q_{1}} \cdot \boldsymbol{q_{2}}
\end{equation}

The final score is the difference between similarity $s$ and threshold $t$:

\begin{equation}
    SCORE = s - t
\end{equation}

If the score is greater than 0, it is a pair of identical commodities, otherwise not. The higher score, the greater probability of being an identical commodity pair.

\subsection{Model Architecture}\label{model}
We use the identical architecture for both streams, but in fact we can design different architectures. Taking threshold stream as example, we have a RoBERTa~\cite{liu2019roberta} to encode textual feature $\boldsymbol{q^u}$ from text $\boldsymbol{u}$ and a Swin Transformer~\cite{liu2021swin} to encode visual feature $\boldsymbol{q^v}$ from image $\boldsymbol{v}$. Then we concatenate them and project the concatenated embedding into a common embedding space by a linear layer $\boldsymbol{h}$:

\begin{equation}
    \boldsymbol{q} = \boldsymbol{h}([\boldsymbol{q}^{\boldsymbol{u}}, \boldsymbol{q}^{\boldsymbol{v}}])    
\end{equation}

Similarly, we can choose other backbones to encode single-modality features.

\begin{figure}
    \centering
    \includegraphics[width=1.0\textwidth]{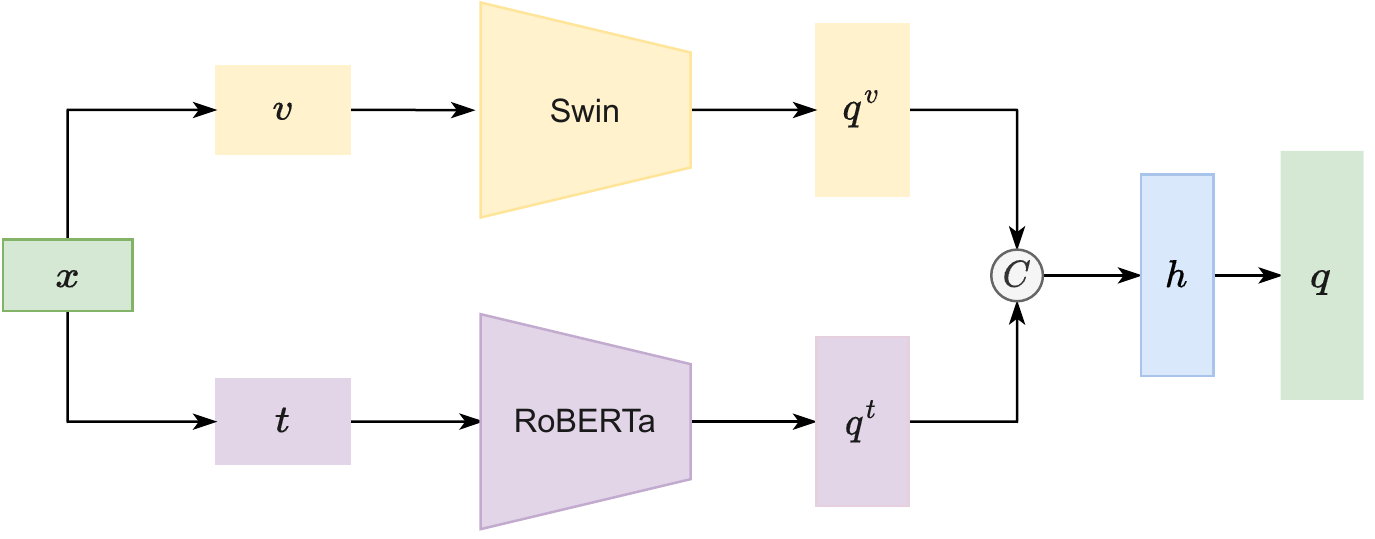}
    \caption{
    Self-adaptive threshold network. We use Swin Transformer~\cite{liu2021swin} and RoBERTa~\cite{liu2019roberta} to encode image features and text features respectively. The features of different modalities are fused by a linear layer.
    }
    \label{fig:fusion}
\end{figure}

\subsection{Loss Function}\label{loss}

We use cross entropy loss~\cite{goodfellow2016deep} to train the model:

\begin{equation}
\mathcal{L}=-\log \frac{\boldsymbol{y}\exp \left(\boldsymbol{p}_{1} \cdot \boldsymbol{p}_{2} \right)+(1-\boldsymbol{y})\exp \left(\boldsymbol{q}_{1} \cdot \boldsymbol{q}_{2} \right)}{\exp \left(\boldsymbol{p}_{1} \cdot \boldsymbol{p}_{2} \right)+\exp \left(\boldsymbol{q}_{1} \cdot \boldsymbol{q}_{2} \right)}\\
\end{equation}

where $\boldsymbol{y}\in\{0, 1\}$ is the ground-truth.

\section{Experiments}

\subsection{Experimental Setup}

\subsubsection{Datasets}

 The official dataset contains about 50,000 commodity pairs for training and about 20,000 commodity pairs for test\footnote{\url{https://tianchi.aliyun.com/competition/entrance/531956/information}}. The training is only conducted on the official training set. We do not use unlabeled data or external dataset during training. When dividing the training and validation sets, we remove the items that appear in the training set to ensure that the training set and validation set do not overlap. The ratio of the final training set and validation set is about 5.6:1. We resize all images to 384 x 384. For text, we take the title and the 10 most frequent attributes as input. We do not apply augmentations on either image or text data.

\subsubsection{Implementation Details}

Our implementation is based on PyTorch~\cite{NEURIPS2019_9015} and HuggingFace~\cite{wolf-etal-2020-transformers}. We initialize the image encoder with Swin Transformer~\cite{liu2021swin}, pre-trained on ImageNet~\cite{deng2009imagenet}. Text encoders are initialized from pre-trained RoBERTa~\cite{liu2019roberta}. We train SAT in an end-to-end manner. For all experiments, we use Adam optimizer~\cite{kingma2014adam} with betas [0.9, 0.999]. We train SAT for 100K steps on 2 NVIDIA A100 GPUs with a total batch size of 8, which takes about 20 hours. The initial learning rate and weight decay are 2e-6 and 1e-6 respectively. We use cosine annealing learning rate decay without warmup.

\subsection{Ablation}

In the ablation study, we validate the effectiveness of our method and analyze the impact of input modalities and pre-trained models. If not mentioned, hyperparameters other than the ablated factor are the same.

\subsubsection{Effectiveness of SAT}

We first build a simple baseline as plotted in the red box of Fig. \ref{fig:model}, which only have a commodity encoder. Besides, we add a Learnable Threshold (LT) to it. The threshold is learnable and the same for all commodities. As shown in Tab. \ref{tab1}, our SAT outperforms baseline methods by a large margin, indicating the effectiveness of SAT. Specifically, SAT brings significant F1-score improvements (i.e. +0.0620 higher than LT).

\begin{table}
    \caption{Results of different methods on the validation set.}\label{tab1}
    \centering
    \begin{tabular}{ccccc}
    \toprule
        Method & F1-score & Precision & Recall & Accuracy \\ 
    \midrule
        Baseline & 0.7250	& 0.6097 & \textbf{0.8940} & 0.6432 \\ 
        LT & 0.8204 & 0.8139 & 0.8270 & 0.8096 \\ 
        SAT & \textbf{0.8824} & \textbf{0.8795} & 0.8853 & \textbf{0.8759} \\ 
	\bottomrule
    \end{tabular}
\end{table}

\subsubsection{Impact of Modality}

We further analyze the input modalities. Tab. \ref{tab2} shows the detailed comparisons. Image-only SAT achieves better performance than text-only, with a lead of 0.0612 on F1 score. Taking text and images together as input can further improve the performance. We believe that SAT can be further enhanced with other modality inputs, which is worth exploring in the future study.

\begin{table}
    \caption{Results of SAT with different input modalities.}\label{tab2}
    \centering
    \begin{tabular}{cccccc}
    \toprule
        Text & Image & F1-score & Precision & Recall & Accuracy \\ 
    \midrule
        \cmark & ~ & 0.7888 & 0.7555 & 0.8251 & 0.7676 \\ 
        ~ & \cmark & 0.8500 & 0.8599 & 0.8403 & 0.8440 \\ 
        \cmark & \cmark & \textbf{0.8824} & \textbf{0.8795} & \textbf{0.8853} & \textbf{0.8759} \\ 
	\bottomrule
    \end{tabular}
\end{table}

\subsubsection{Impact of Pre-trained Models.}

We also study the impact of pre-trained models. As mentioned above, we use Swin Transformer~\cite{liu2021swin} pre-trained on ImageNet-1k and ImageNet-22k and pre-trained RoBERTa~\cite{liu2019roberta}. In this ablation, we random initialize the Swin Transformer~\cite{liu2021swin} and RoBERTa~\cite{liu2019roberta}. As shown in Tab. \ref{tab3}, we observe significant performance improvement with pre-trained models, which indicates the importance of pre-trained models. 

\begin{table}[t]
    \caption{Ablation study of pre-trained models}\label{tab3}
    \centering
    \begin{tabular}{cccccc}
    \toprule
        Pre-trained & F1-score & Precision & Recall & Accuracy \\ 
    \midrule
        \xmark & 0.7815 & 0.7606 & 0.8037 & 0.7637 \\ 
        \cmark & \textbf{0.8824} & \textbf{0.8795} & \textbf{0.8853} & \textbf{0.8759} \\ 
	\bottomrule
    \end{tabular}
\end{table}

\subsection{Score Distribution}\label{distribution}
\begin{figure} 
  \centering 
  \subfigure[LT\label{fig:lt-distribution}]{ 
    \includegraphics[width=0.45\textwidth]{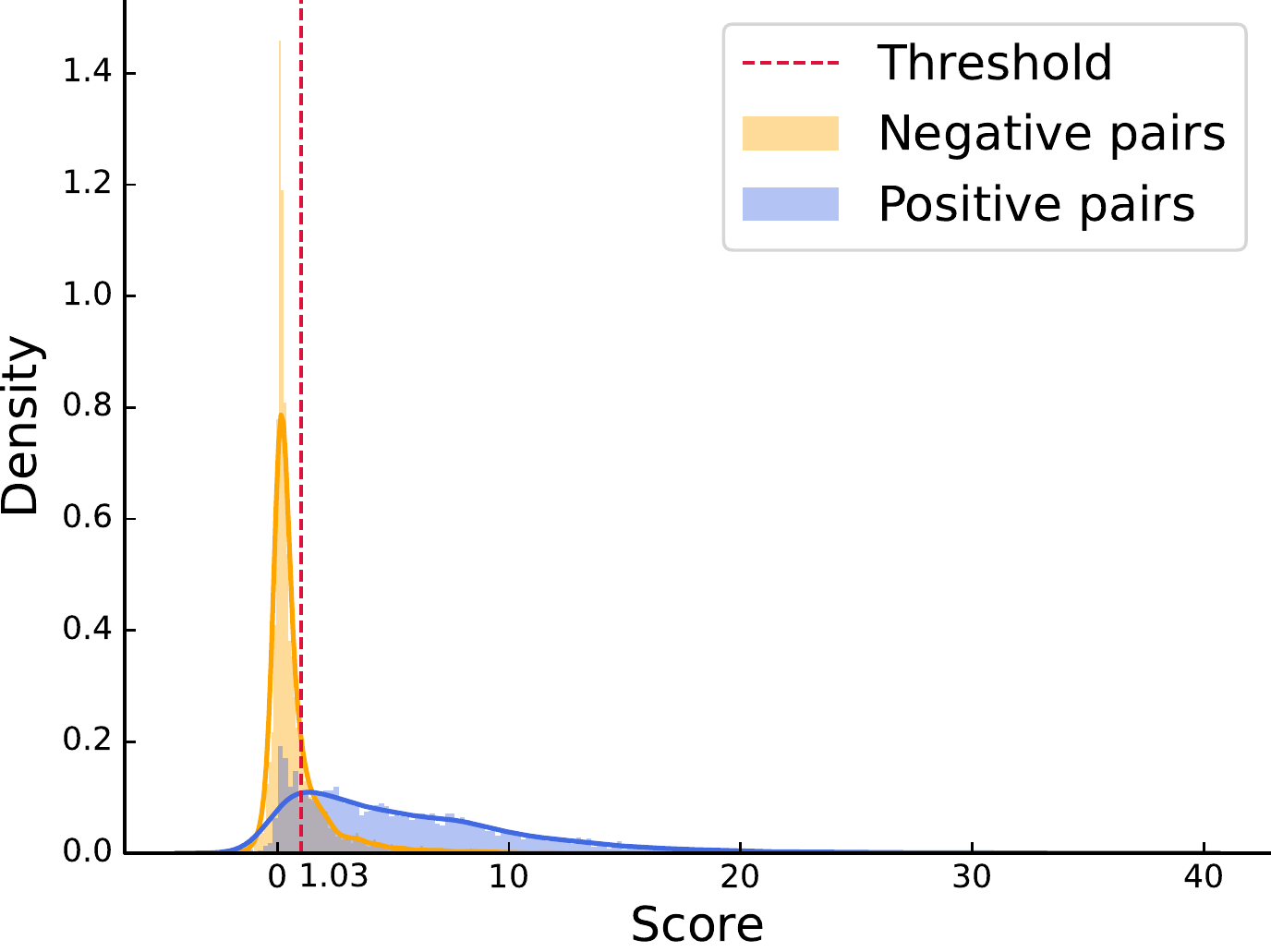} 
  } 
  \subfigure[SAT\label{fig:sat-distribution}]{ 
    \includegraphics[width=0.45\textwidth]{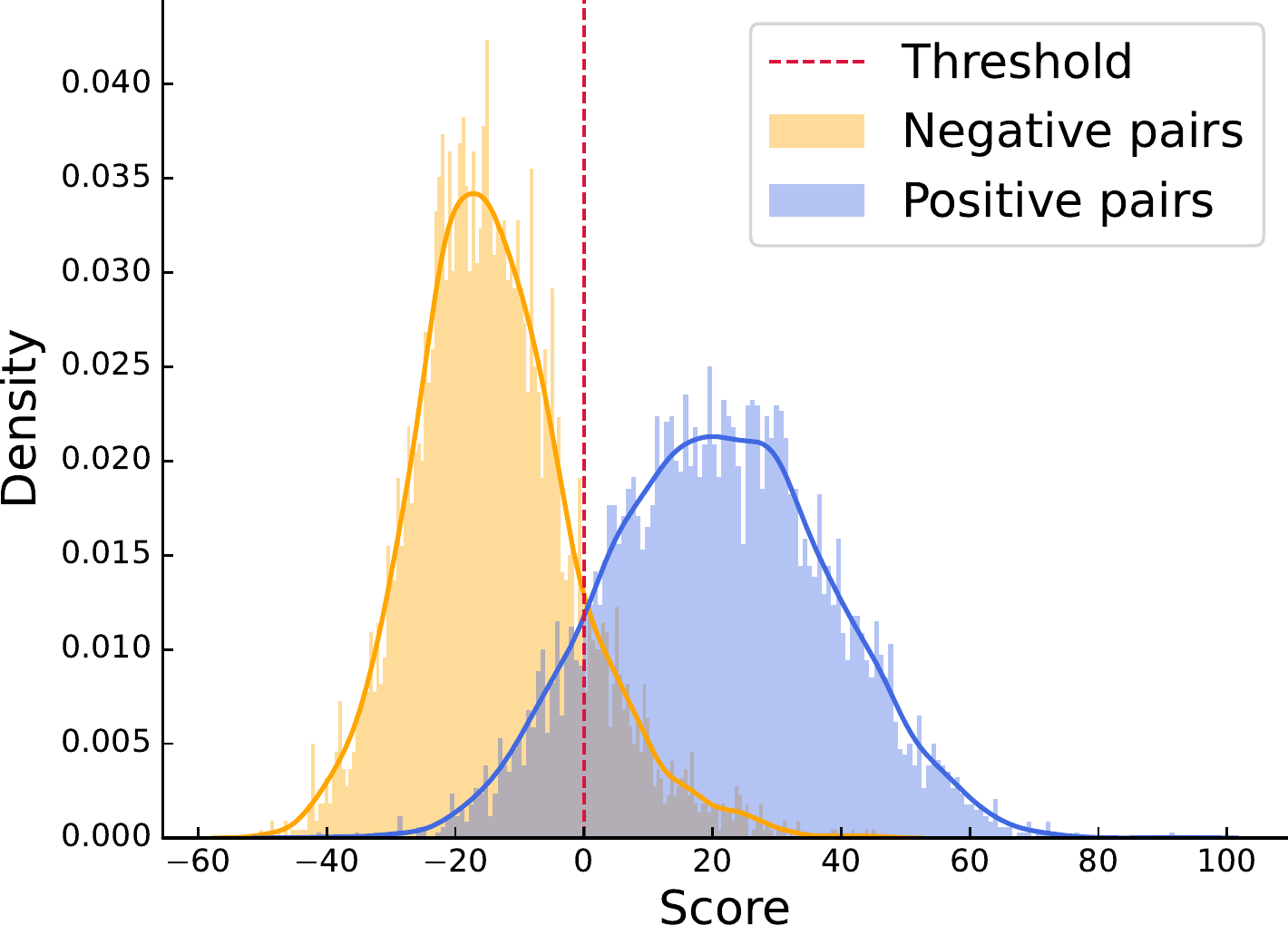} 
  } 
  \caption{\textbf{Visualization of score distribution.} We show histograms of scores of LT and SAT. The density of the score is estimated by kernel density estimation. Compared to LT, the peak density of SAT is lower and farther away from the threshold. } 
  \label{fig:distribution}
\end{figure}

Fig. \ref{fig:distribution} shows the score distribution of LT and SAT. As shown in Fig. \ref{fig:lt-distribution}, the density peak of negative pairs is high with LT. In the meanwhile, the density peak of both positive and negative pairs is near the threshold, which means there are quantities of pairs around the threshold. The higher density peak and the closer density peak to the threshold, the more susceptible to threshold changes. A slight fluctuation of the threshold may affect the performance much. In addition, LT heavily depends on the training data distribution, not conducive to model generalization. By contrast, the density curve of SAT is much smoother and much easier to more distinctive than LT as shown in Fig. \ref{fig:lt-distribution}. It can be seen that the density peak of SAT is lower and farther away from the threshold. This indicates that default threshold 0 is virtually an optimum. Therefore without manually adjusting the threshold, we can distinguish positive and negative pairs by the default threshold of 0.

\section{Conclusion}
In this paper, we first analyzed the potential problems of traditional representation learning in the commodity verification task. Then we proposed SAT and demonstrated its effectiveness and advantages by quantitative experiments and score distribution visualization. With SAT, we obtained a representative and discriminative commodity representation space and achieved excellent performance. As future work, we would like to extend SAT to other multimodal representation learning tasks. 

%
%

%
%
%
\bibliographystyle{splncs04}
\bibliography{mybibliography}
%





\end{document}